\documentclass[conference]{IEEEtran}
\usepackage{graphicx}
\usepackage{subfigure}
\usepackage{amsmath}
\usepackage{amssymb}
\usepackage{bm}
\usepackage{multirow}
\usepackage{indentfirst}
\usepackage{mathrsfs}
\usepackage{url}
\urldef{\mailsa}\path||
\urldef{\mailsb}\path||
\ifCLASSINFOpdf
\else
\fi

\begin{document}
%
\title{GetNet: Get Target Area for Image Pairing}


\author{\IEEEauthorblockN{Henry H. Yu}
\IEEEauthorblockA{Department of Automation\\Tsinghua University\\
Beijing, China\\
Email: afishyay@gmail.com}
\and
\IEEEauthorblockN{Jiang Liu}
\IEEEauthorblockA{Department of Automation\\
Tsinghua University\\
Beijing, China\\
Email: liujiang15@mails.tsinghua.edu.cn}
\and
\IEEEauthorblockN{Hao Sun}
\IEEEauthorblockA{Viterbi School of Engineering\\
University of Southern California\\
Los Angeles, USA\\
Email: sh1005536292@163.com}
\and
\IEEEauthorblockN{Ziwen Wang}
\IEEEauthorblockA{Metropolitan College\\
Boston University\\
Boston, USA\\
Email: asoapqaq@bu.edu}
\and
\IEEEauthorblockN{Haotian Zhang}
\IEEEauthorblockA{College of Bioinformatics Science and Technology\\
Harbin Medical University\\
Harbin, China\\
Email: zhang@haotian.info}}


%




\maketitle

\begin{abstract}
Image pairing is an important research task in the field of computer vision. And finding image pairs containing objects of the same category is the basis of many tasks such as tracking and person re-identification, etc., and it is also the focus of our research. Existing traditional methods and deep learning-based methods have some degree of defects in speed or accuracy. In this paper, we made improvements on the Siamese network\cite{melekhov2016siamese} and proposed GetNet. The proposed method GetNet combines STN\cite{jaderberg2015spatial} and Siamese network to get the target area first and then perform subsequent processing. Experiments show that our method achieves competitive results in speed and accuracy.

\end{abstract}


%
\IEEEpeerreviewmaketitle

\section{Introduction}
Find image pairs with a certain connection is a basic technology in the field of computer vision. The essence of many research fields is the process of finding image pairs, such as image matching, image retrieval, etc. In some other research areas, image pairing also plays a key role, such as tracking, object recognition, multi-view 3D reconstruction, structure-from-motion (SfM) and so on. The problem we are concerned here is whether the image contains objects of the same category, which is very common in tasks such as image matching, image retrieval and tracking. Although the rapid development of deep learning method in recent years has greatly promoted the advancement of computer vision and related fields, finding image pairs that meet certain criteria across large, unstructured image datasets can be very time-consuming and prone to errors, especially when the target object is small in the image or it is in a cluttered background. 

Recently, Siamese architecture\cite{melekhov2016siamese} has been utilised in various image pairing problem, such as face verification, local image patches pairing as well as whole-image matching, but not yet in generic object-centred image pairing and retrieval. In this paper, we establish a theoretical connection between Spatial Transformer Networks (STNs) and Siamese networks, which can find matching or non-matching image pairs (i.e. image pairs that contains the same object or not) as well as output the common parts of matching pairs (i.e. target objects) since STNs can apply the affine transformation to the images which can help extract the region of interest (ROI). An architectural network is designed that is trained by inputting example images pairs and  supervised by simply labelled "1"(means pairing) or "0"(means non-pairing). A new way of network training was also proposed to get better performance. Our experimental results show that the proposed method improves pairing performance compared to the original Siamese networks. With regions of interest output by STNs, our method also provides a convenient way for locating valuable part in image and an effective way of dataset ground truth auto-labelling.

To sum up, the main contributions of our work are:
\begin{description}
  \item[\romannumeral1]  Propose a method for object-oriented image pairing that can extract specific subparts that are relevant for image retrieval and matching in clutter with just 0/1 supervision.
  \item[\romannumeral2] Provide a convenient way for locating valuable part in image and an effective way of dataset ground truth auto-labelling.
  \item[\romannumeral3] Propose a new way to train the network to get better performance.
\end{description}

The paper is organized as follows. We first introduce the related work in Section  \uppercase\expandafter{\romannumeral2} and then give a general review of Spatial Transformer Networks and Siamese and describe our proposed network structures in Section \uppercase\expandafter{\romannumeral3}. The details of evaluation datasets and experimental results are given in Section  \uppercase\expandafter{\romannumeral4} and the future work is in Section \uppercase\expandafter{\romannumeral5}.
Finally, our conclusion is presented in Section \uppercase\expandafter{\romannumeral6}.
 
\section{Related Work}
\bfseries Image matching \mdseries
 is an important problem in computer vision and it can be seen as a sub-question of the image pairing problem. Although recent rapid advances in convolutional neural networks (CNNs) techniques have achieved state-of-the-art performance in tasks such as image recognition, object segmentation and so on, finding similar images (e.g. images that have the same objects) across a large, unstructured image dataset can still be very time-consuming, and the performance will be worse if the object in the image is relatively small or the image has a cluttered background, which remains a common problem in computer vision.
The main obstacles toward image matching include viewpoint variation, scale variation, illumination variation, occlusion and background clutter. Over the years, different methods have been proposed to solve the image matching problem and increase the accuracy and performance. Generally, these methods can be split into two categories. The first category is based on hand-crafted image feature extraction. Jyoti $et\; al$. used SIFT feature to match stereo image pair that be applied 3D reconstruction\cite{joglekar2010image}. Also, SURF feature\cite{pang2012fully}, ORB feature\cite{li2016improved}, color histogram\cite{zhang2014learning}and HOG feature can be used to do the image matching in order to reduce the time of computation but still need lots of time. The researchers also proposed many other methods for extracting features or key points to do the matching job like CSIFT\cite{burghouts2009performance}, BRISK\cite{leutenegger2011brisk}, ORB\cite{rublee2011orb}, FREAK\cite{alahi2012freak}, stereo keypoint matching\cite{seabright2018simple}, LBP\cite{ahonen2006face} and so on. The second category is based on convolutional neural networks (CNNs). CNNs have been widely applied into many areas in computer vision including image matching and made remarkable performances better than traditional methods. Iaroslav $et\; al$.\cite{melekhov2016siamese} presented a method to measure the whole-image similarity based on deep neural network and predict the similarity of a query image pair, showing very promising results. However, this method needs to use a pre-trained CNN classifier which causes inconvenience. And since it adopts whole-image similarity to measure the similarity of the image pair, it cannot perform very well in situations where the similar parts are relatively small in the image pairs or the backgrounds are cluttered. Although many improved networks based on Siamese such as SConE\cite{trzcinski2018scone}, Patch Match Networks\cite{hanif2019patch}, SimNet\cite{appalaraju2017image} and some other deep convolutional neural networks based methods like \cite{kumar2018deep}, \cite{weinzaepfel2013deepflow} are proposed, it is still far from resolved.

\bfseries Image retrieval \mdseries
has become an important research area in computer vision these years. Its task is to find images that have some connection with the query image, whose essence is actually image pairing. Image retrieval is classified mainly in several types such as text based, content based, sketch based and so on. Here our focus is on content based image retrieval (CBIR) since it retrieval image based on the content which is similar to our research. In CBIR, different researchers focus on different aspects and have achieved good results. Chang $et\; al$\cite{chang2005color} proposed the image retrieval using the color distribution, mean and the standard deviation and Sun $et\; al$\cite{sun2006image} suggested a color distribution entropy method. There are also some researchers see shape as an important feature\cite{liu2007survey}\cite{stehling2000shapes} and some tend to texture\cite{kekre2012image}\cite{sandhu2012content}. 
In addition, the kernel-based approach proposed by Karmakar $et\; al$ \cite{karmakar2018kernel} is also very instructive. 
And just like the image matching, the research of algorithms in image retrieval can be divided into traditional methods and deep learning based methods. In the former aspect, Krishna $et\; al$ \cite{raja2013content} proposed an indexing of the image using the k-means algorithm and Sonali $et\; al$\cite{jain2013novel} proposed the SVM algorithm to act as a classifier. Also, the success of deep neural networks on feature representation has led it be widely used in image retrieval tasks. Models pre-trained on popular datasets such as ImageNet\cite{deng2009imagenet}, Landmarks\cite{babenko2014neural}, COCO\cite{lin2014microsoft}, etc. can be used to extract features of images and are found to have good generalization performance. Especially, convolutional layers have been proved to be
most beneficial at retrieving images \cite{babenko2014neural}\cite{gordo2016deep}\cite{radenovic2016cnn}\cite{razavian2016visual}. And then, nearest neighbor search is used on the feature vectors to find the most similar images to a query. Although such progress has been made, speed and accuracy are still problems that need to be solved during large-scale retrieval.



\section{PROPOSED METHOD}
\subsection{Spatial Transformer Networks}
In our method, Spatial Transformer Networks (STNs)\cite{jaderberg2015spatial}as Fig 1 is applied to process the two images of every pair separately. The spatial transformer consists of three parts including localisation net, grid generator and sampler. First, the localisation net takes the original image $U\in H \times W \times C$ where $H$ is the height, $W$ is the width and $C$ is the number of channels as input and output the parameters $\theta_{ij}$ which are related to the transformation
\begin{equation}
 A_\theta = {\left[ \begin{array}{ccc}
\theta_{11} & \theta_{12} & \theta_{13} \\
\theta_{21} & \theta_{22} & \theta_{23} \\
\end{array}
\right ]}
\label{eq_1}
\end{equation}
Second, grid generator generates the parameterized sampling gird and then, by applying the grid to the original input image, deformed output image $V \in H' \times W'\times C'$  with height $H'$, width $W'$ and channels $C'$ is produced.

To apply the sampling gird into the input image, all output pixels that are defined on a regular gird $G$ with coordinate $(x_i^t, y_i^t)$ are computed to form the output image. And since localisation net of 2D affine transformation can output six parameters and it means that STN can apply an affine transformation to the original image like below.
\begin{equation}
\left(\begin{array}{c}
x_i^s\\
y_i^s\\
\end{array} \right) = A_{\theta} \left(\begin{array}{c}
x_i^t\\
y_i^t\\
1\\
\end{array} \right)= {\left[ \begin{array}{ccc}
\theta_{11} & \theta_{12} & \theta_{13} \\
\theta_{21} & \theta_{22} & \theta_{23} \\
\end{array}
\right ]}\left(\begin{array}{c}
x_i^t\\
y_i^t\\
1\\
\end{array} \right)
\label{eq_2}
\end{equation}
Where $x_i^t$ and $y_i^t$ are the coordinates in the output images and $x_i^s$ and $y_i^s$ are coordinates in the input feature maps. $A_{\theta}$ represents affine transformation. Thus, the precise image which contains the target object from the original image can be extracted using STN.

In our experiment, in order to facilitate the training process, the network is modified and the localisation net only output three parameters including $s$, $t_x$ and $t_y$ which can achieve the local translation in the original images as below:
\begin{equation}
 A_\theta = {\left[ \begin{array}{ccc}
s & 0 & t_x \\
0 & s & t_y \\
\end{array}
\right ]}
\label{eq_3}
\end{equation}
Where $t_x$ represents the distance of translation in the x axis. $t_y$ represents the distance of translation in the $y$ axis and $s$ represents the cropping ratio. And when the $x_i^s$ and $y_i^s$ which define the spatial location in the input feature maps are obtained, the output feature maps can be calculated as below:
\begin{equation}
\begin{split}
 V_i^c=\sum_{n}^{H}\sum_{m}^{W}U_{nm}^ck(x_i^s-m;\Phi_x)k(y_i^s-n;\Phi_y)\\
 \forall i\in {\left[ \begin{array}{ccc}
1 & \cdots & H'W' \\
\end{array}
\right ]}\;\;\forall C\in {\left[ \begin{array}{ccc}
1 & \cdots & C \\
\end{array}
\right ]}
\label{eq_4}
\end{split}
\end{equation}
Where $k()$ is the image interpolation kernel function (e.g. bilinear, nearest neighbor and so on ) and $\Phi_x$ and $\Phi_y$ are the parameters of $k()$. $U_{nm}^c$ is the value at location $(n,m)$ in the input feature map in channel $c$ and $V_i^c$ is the value at location $(x_i^s,y_i^s)$ in the output feature map in channel $c$. And in order to allow backpropagation, the sampling kernel can be used only when gradients can be defined about $x_i^s$ and $y_i^s$. Take bilinear sampling kernel given as below as an example,

\begin{equation}
 V_i^c=\sum_{n}^{H}\sum_{m}^{W}U_{nm}^c \textrm{max}(0,1-|x_i^s-m|)\textrm{max}(0,1-|y_i^s-n|)
\label{eq_5}
\end{equation}
the gradients with respect to $U$ and $G$ of it can be defined and the partial derivatives are as below:

\begin{equation}
 \frac{\partial V_i^c}{\partial U_{nm}^{c}}=\sum_{n}^{H}\sum_{m}^{W}\textrm{max}(0,1-|x_i^s-m|)\textrm{max}(0,1-|y_i^s-n|)
\label{eq_6}
\end{equation}

\begin{equation}
 \frac{\partial V_i^c}{\partial x_i^s}=\sum_{n}^{H}\sum_{m}^{W}U_{nm}^c \textrm{max}(0,1-|y_i^s-n|)\left\{ \begin{array}{ll}
0 & \textrm{if $|m-x_i^s|\geq 1$}\\
1 & \textrm{if $m \geq x_i^s$}\\
-1 & \textrm{if $m < x_i^s$}
\end{array} \right.
\label{eq_7}
\end{equation}
and the same for $ \frac{\partial V_i^c}{\partial y_i^s}$.
So the STNs can achieve end-to-end train and exact ROI to carry out subsequent processing.
\begin{figure}[tb]
  \centering
  \centerline{\includegraphics[width=7cm]{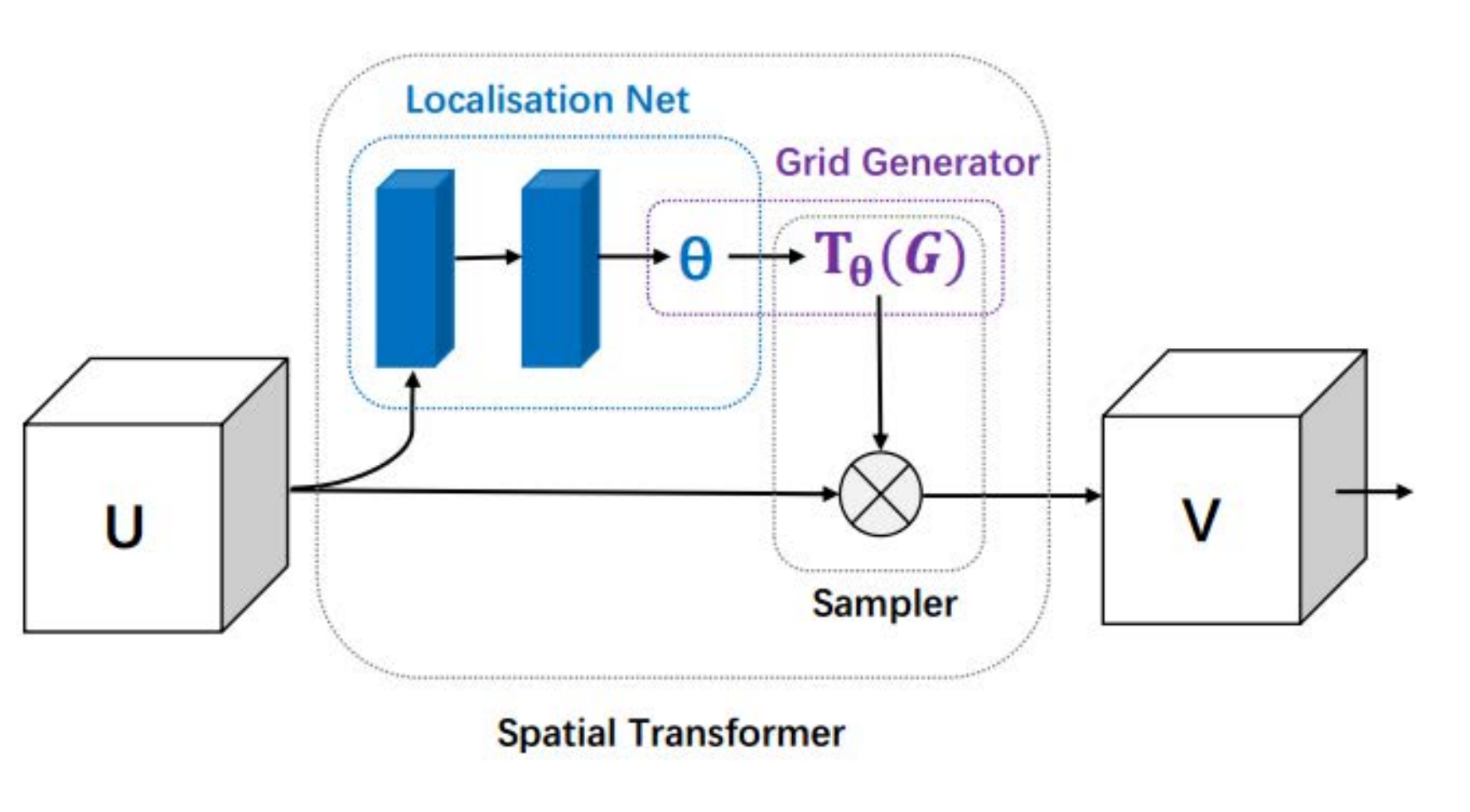}}
\caption{Spatial Transformer Networks.}
\label{fig:Fig.1}
\end{figure}

\subsection{Siamese CNN architecture}
A Siamese CNN architecture is used to match the image pair as Fig 2. Siamese CNN architecture is a classical algorithm which first extracts features from input pair and then compares the features to calculate the similarity of the input pair. The detail of the neural networks is as below and the two one-dimensional feature vectors exacted are connected into a one-dimensional feature vector and then input into the fc layer.

The contrastive loss \cite{chopra2005learning} (which is defined as below) of the output features is applied to measure the similarity of the images in every pair and carry out a simple supervised learning by giving label 1 or 0 to indicate whether the pairing is successful.
\begin{equation}
 L=\frac{1}{2N}\sum_{n=1}^{N}yd^2+(1-y)\textrm{max}(margin-d,0)^2
\label{eq_8}
\end{equation}
Where d is the Euclidean distance \cite{Ghafoor2003Image} of the features of image pair as below and y is the label 1 or 0. Margin is a given threshold.
\begin{equation}
 d=||a_n - b_n||_2
\label{eq_9}
\end{equation}

\begin{figure}[tb]
  \centering
  \centerline{\includegraphics[width=8cm]{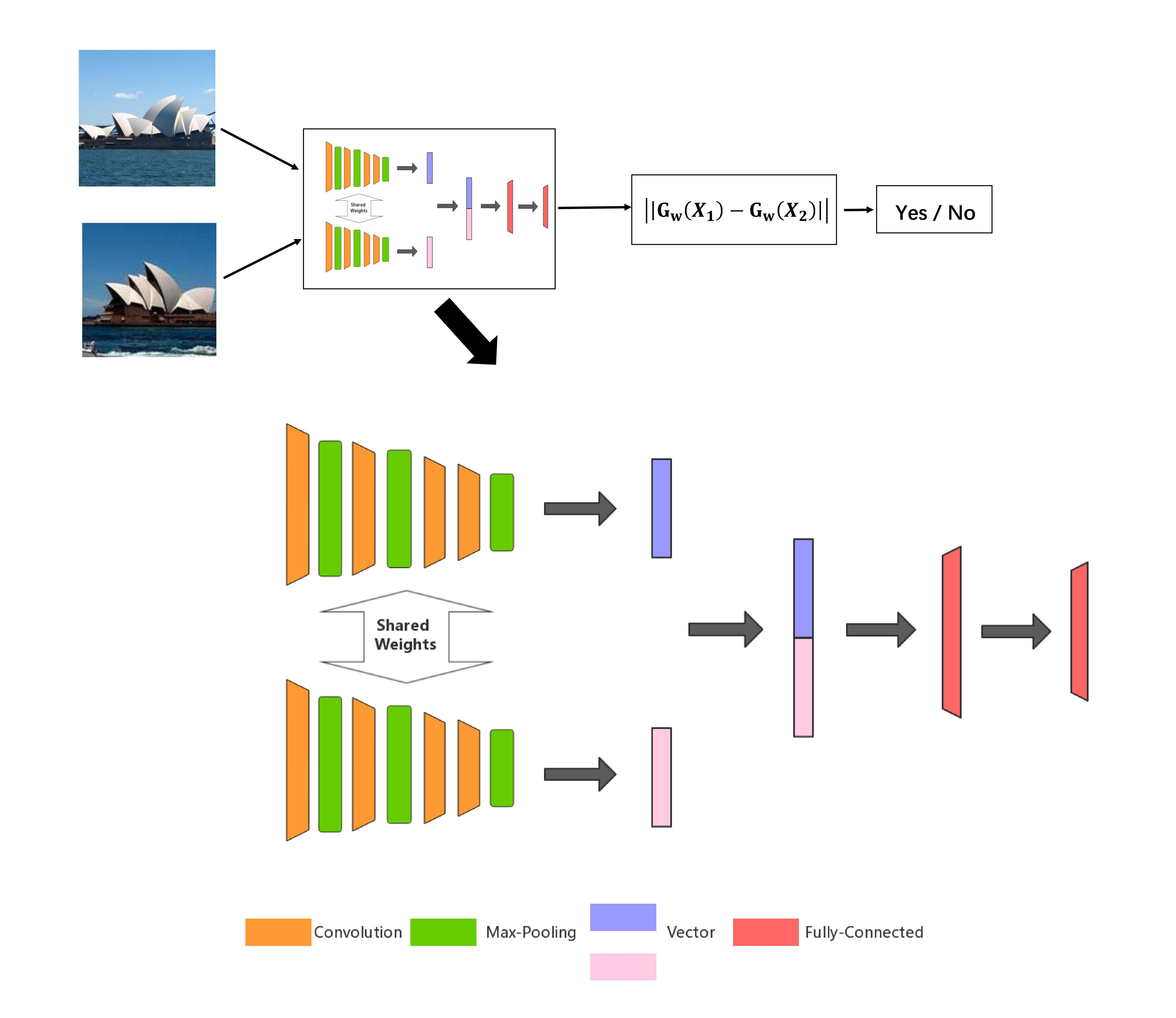}}
\caption{Siamese CNN architecture.}
\label{fig:Fig.2}
\end{figure}

\subsection{Proposed network}
\begin{figure}[tb]
  \centering
  \centerline{\includegraphics[width=9cm]{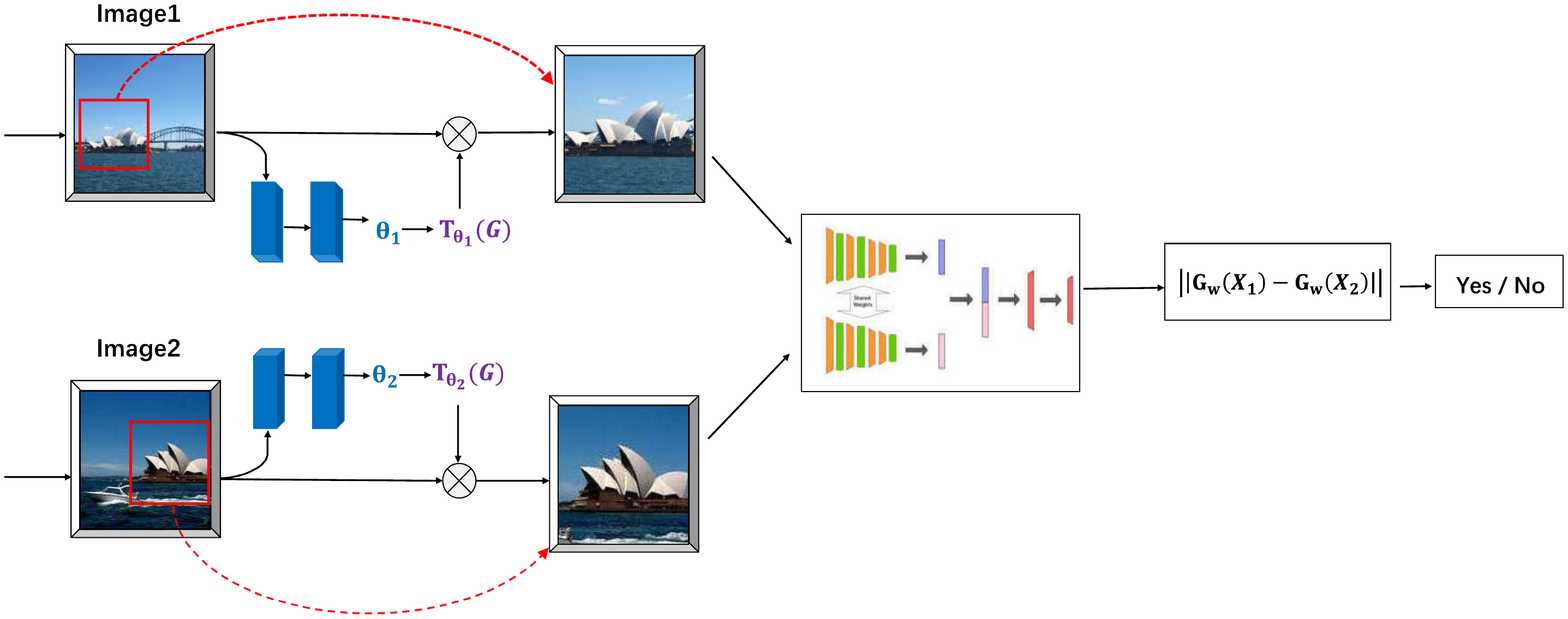}}
\caption{Structure of proposed network.}
\label{fig:Fig.3}
\end{figure}

The original image pair is the input of the STN networks and the Siamese networks which share parameters to extract features take the output image pair of STN networks as the input pair. Then the Siamese networks output the predict results using contrastive loss. The proposed network is named GetNet and is shown as Fig.~\ref{fig:Fig.3}.

\subsection{Network training}
A new approach is proposed here to train the GetNet network. CNNs use the back-propagation propagation algorithm to update the gradient for training. But traditional end-to-end training method has less impact on the front end of the network, especially a relatively weak way of supervision only with label 1 or 0 is used here, which makes it even more difficult to update the STN. So a strategy of training the STN and the overall network alternately is proposed. When freezing Siamese network part of the parameters and training the STN alone, the parameters of STN can be adjusted adequately and sample the target object more accurately from the input image. And when training the overall networks, the Siamese network can extract more suitable features to test the similarity according to the label.

Here these two kinds of ways are used in turn and achieved good results which demonstrates the effectiveness of this training approach.

\section{EXPERIMENTAL RESULTS}

\subsection{Dataset}

\subsubsection{MNIST}
\begin{figure}
\centering
\subfigure[]{
\includegraphics[height=0.1\textwidth]{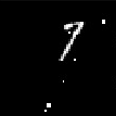}}
\subfigure[]{
\includegraphics[height=0.1\textwidth]{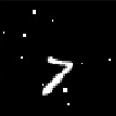}}
\subfigure[]{
\includegraphics[height=0.1\textwidth]{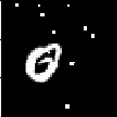}}
\subfigure[]{
\includegraphics[height=0.1\textwidth]{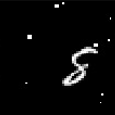}}
\caption{A $60$ pixels $ \times 60$ pixels background with random noises.}
\label{fig:Fig.4}
\end{figure}

MNIST \cite{lecun1998gradient} is a dataset of handwritten digits and the size of all the MNIST original images is $28$ pixels $ \times 28$ pixels. To test our networks, a ``distorted MINIST" is made by putting the images from MINIST dataset into a $60$ pixels $ \times 60$ pixels background and add random noises like Fig.~\ref{fig:Fig.4}.

Apparently that distorted MNIST dataset is more difficult to pair than MNIST and it is suitable to test the performance of our network.
\subsubsection{``Shelf $ \& $ Tote" Benchmark Dataset}
The Shelf $\&$ Tote Benchmark Dataset\cite{zeng2016multi} was created by team MIT and Princeton Vision Group for the worldwide Amazon Picking Challenge 2016 which contains $452$ scenes with $2087$ unique object poses seen from multiple viewpoints.

\begin{figure}
\centering
\subfigure[]{
\includegraphics[height=0.08\textwidth]{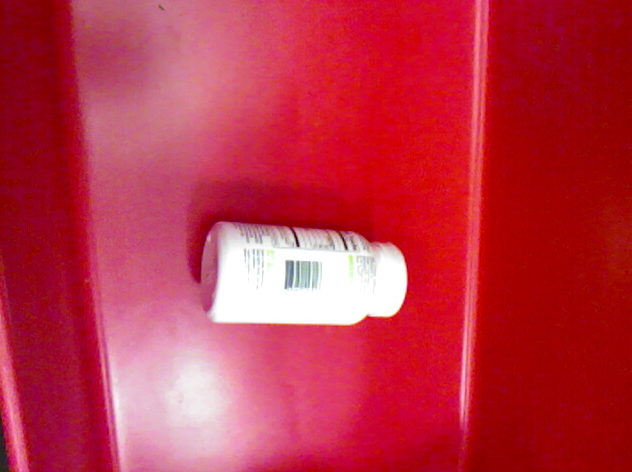}}
\subfigure[]{
\includegraphics[height=0.08\textwidth]{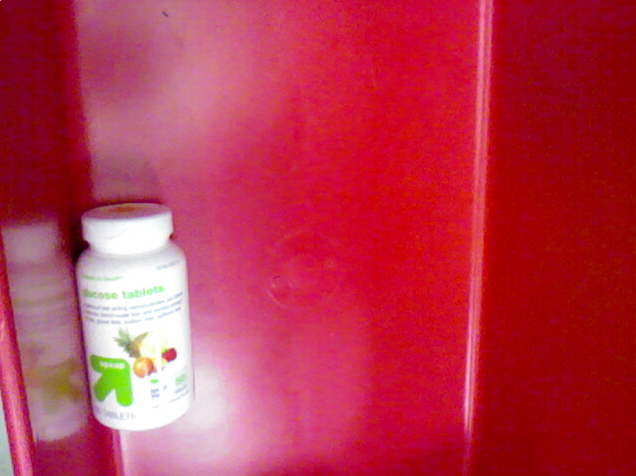}}
\subfigure[]{
\includegraphics[height=0.08\textwidth]{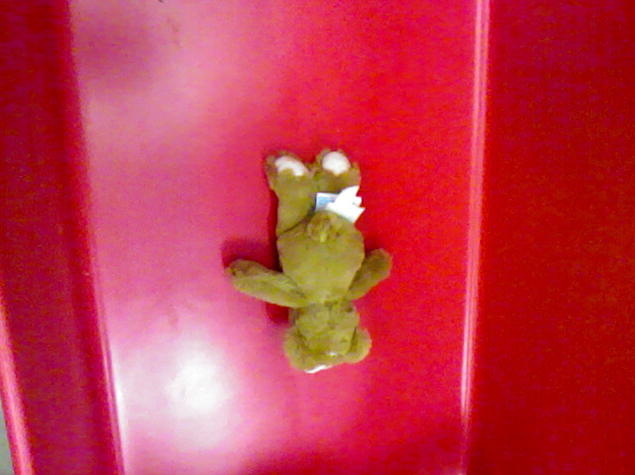}}
\subfigure[]{
\includegraphics[height=0.08\textwidth]{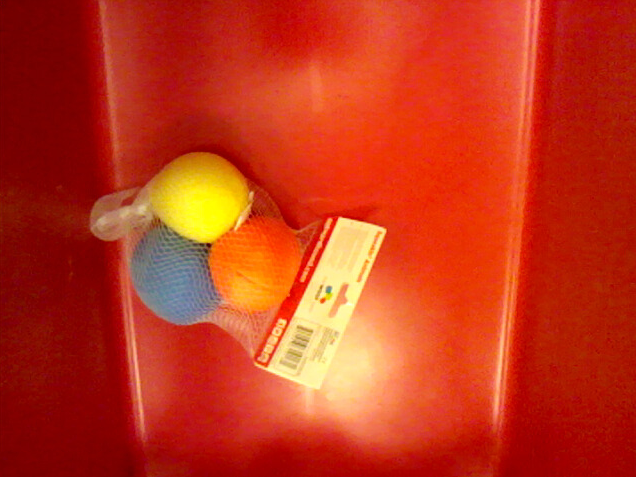}}
\caption{The Shelf \& Tote Benchmark Dataset.}
\label{fig:Fig.5}
\end{figure}

It was used to do self-supervised deep Learning for 6D pose estimation in the Amazon Picking Challenge and here we found it qualified for evaluating the performance of our network. The dataset images are like Fig.~\ref{fig:Fig.5}.

$10$ kinds of objects were picked from the dataset and all the images were reshaped to $160$ pixels $ \times 160$ pixels to do the pairing performance test of the network.
\subsubsection{Caltech Leaves Dataset}
\begin{figure}
\centering
\subfigure[]{
\includegraphics[height=0.07\textwidth]{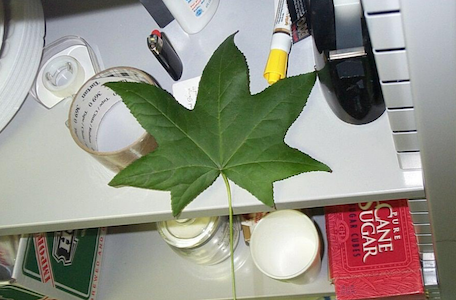}}
\subfigure[]{
\includegraphics[height=0.07\textwidth]{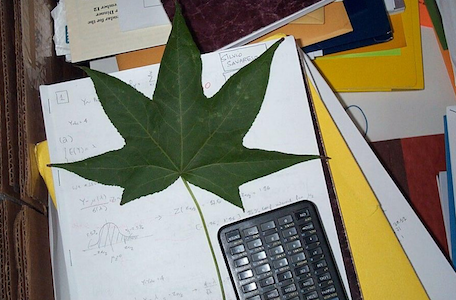}}
\subfigure[]{
\includegraphics[height=0.07\textwidth]{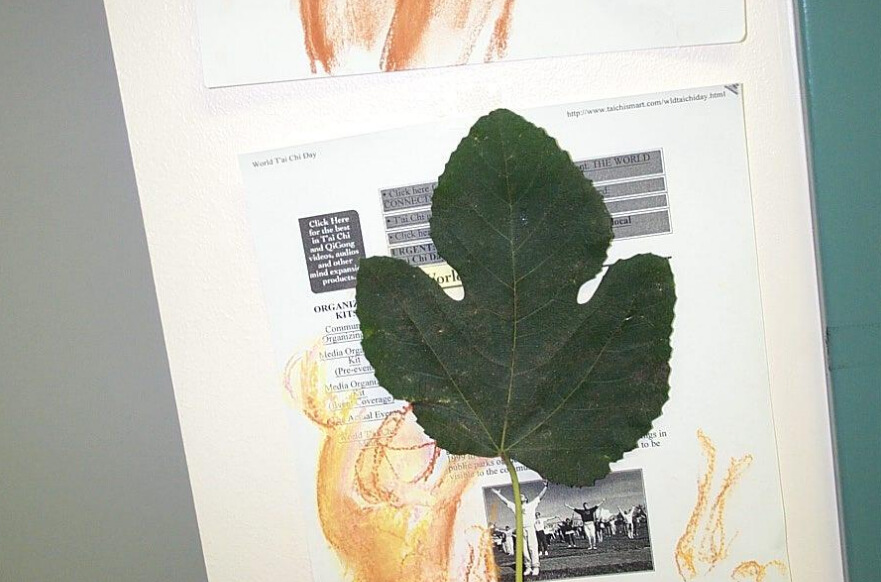}}
\subfigure[]{
\includegraphics[height=0.07\textwidth]{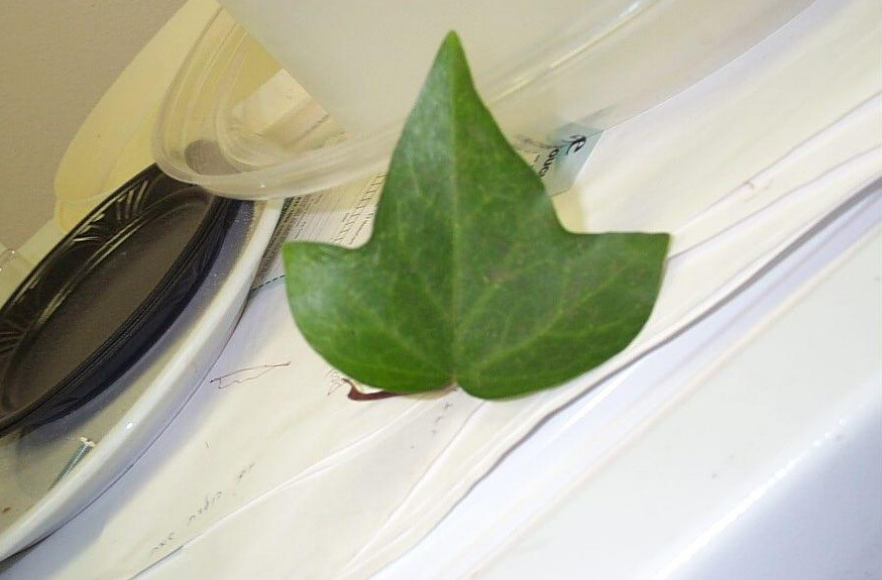}}
\caption{The Caltech leaves dataset.}
\label{fig:Fig.6}
\end{figure}

Caltech leaves dataset is a dataset in Caltech computational vision and it contains $186$ images of $3$ species of leaves against different backgrounds like Fig.~\ref{fig:Fig.6}.

The images in this dataset were also reshaped to $160$ pixels $ \times 160$ pixels and used to do the experiment.
\subsection{Performance}
The three dataset mentioned before was used and for each image in each dataset, another image from the same category is taken to create one pair and is given label 1 and also take an image from a different category to create one pair and is given label 0. Thus three datasets only with label $1$ or $0$ are obtained and the number of positive samples and the number of negative samples in them is nearly equal. And the traditional Siamese networks are used as a comparison to evaluate our network performance.

\begin{table}
\centering
\caption{the result of the experiments}
\scalebox{1.3}{
\begin{tabular}{|c|c|c|}
\hline
       & Siamese (\%) & Our network (\%)\\
 \hline
 MNIST & 98.2 & 99.3\\
 \hline
 Tote dataset & 80.4 & 87.1\\
 \hline
 Leaves dataset & 84.3 & 88.6\\
 \hline
\end{tabular}}
\label{tab:t1}
\end{table}

\begin{figure}[tb]
  \centering
  \centerline{\includegraphics[width=6cm]{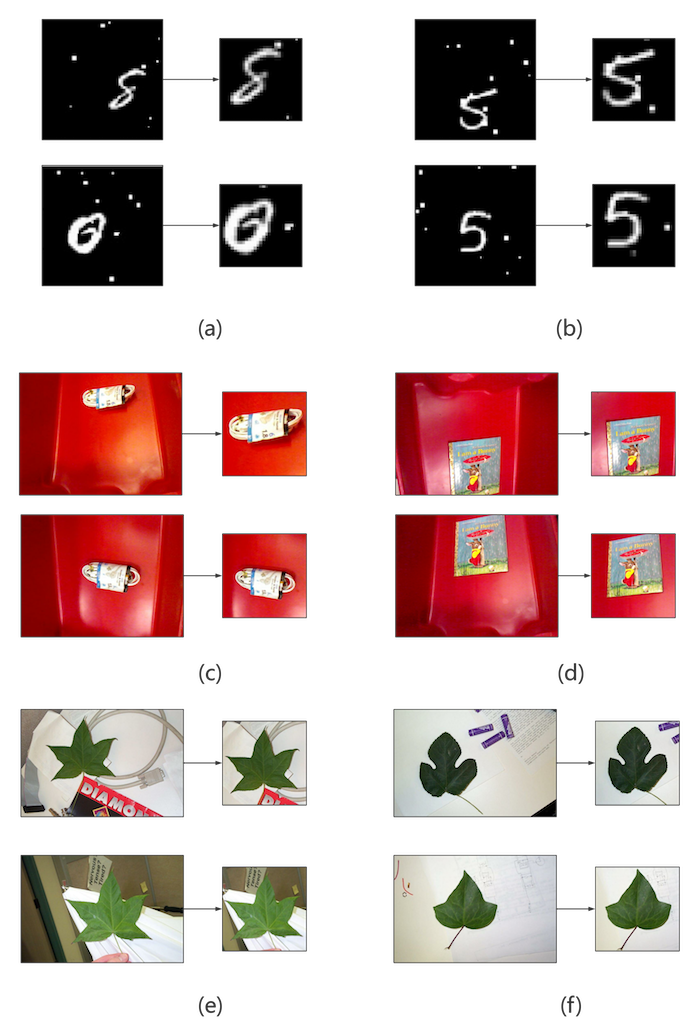}}
\caption{The results of our network performance.}
\label{fig:Fig.7}
\end{figure}

Table~\ref{tab:t1} is the result of the experiments and some of the results are as Fig.~\ref{fig:Fig.7}. 
The left pairs are input image pairs and the right pairs are STN output image pairs. It can be seen that the output images contain the target object more precisely and it is obvious that Siamese network can perform better using the right image pairs. And also it can be seen from above that a very simple label is used to supervise the training procedure, and STN network can still locate the target object of the input image commendably which completes the label of target object in the original image, so it can also eliminate the manual process of data labeling in some special tasks.

\begin{figure}
\centering
\subfigure[]{
\includegraphics[height=0.2\textwidth]{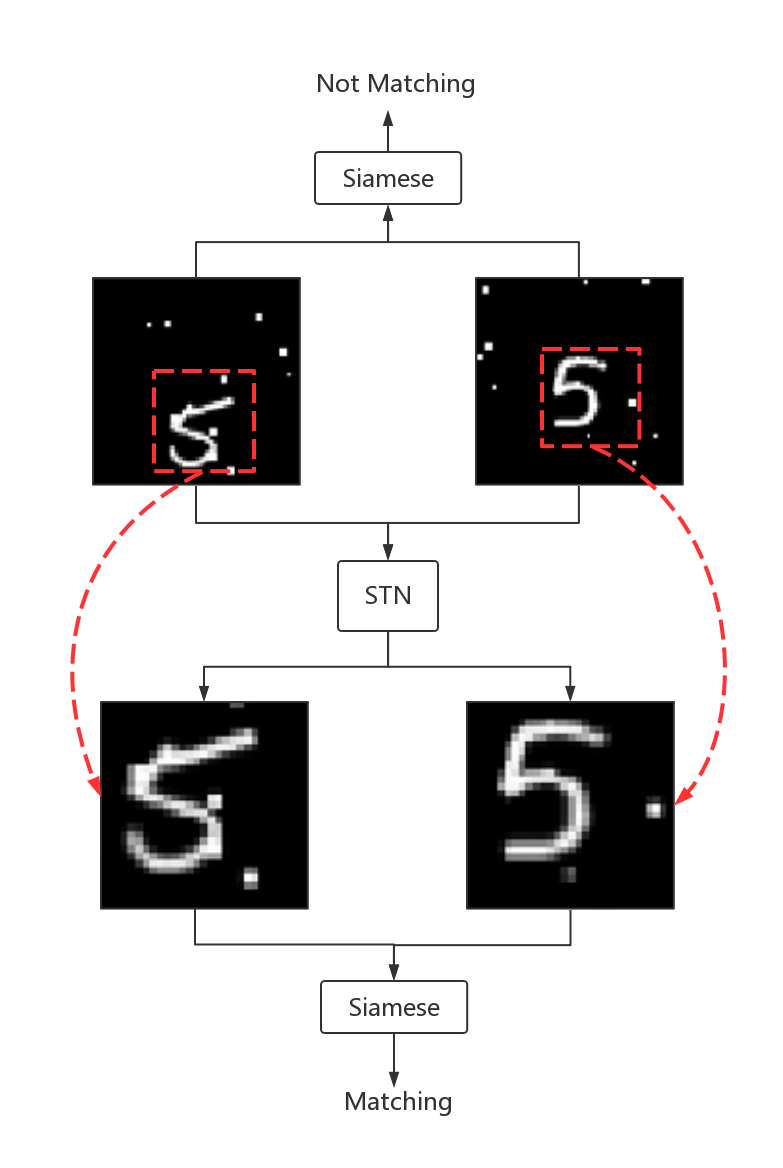}}
\subfigure[]{
\includegraphics[height=0.2\textwidth]{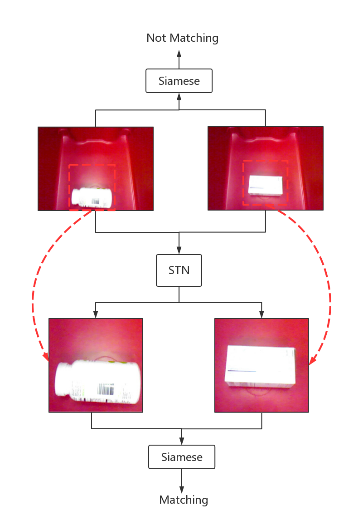}}
\subfigure[]{
\includegraphics[height=0.2\textwidth]{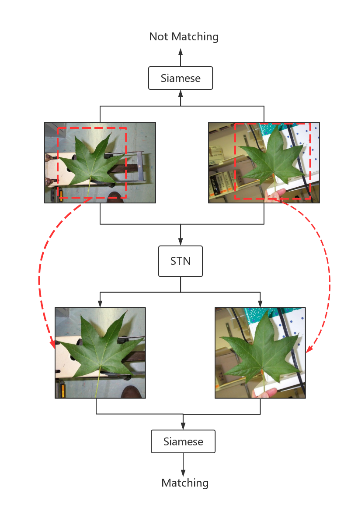}}
\caption{Some examples that traditional Siamese network failed but our network succeeded.}
\label{fig:Fig.8}
\end{figure}

\begin{figure}
\centering
\subfigure[]{
\includegraphics[height=0.2\textwidth]{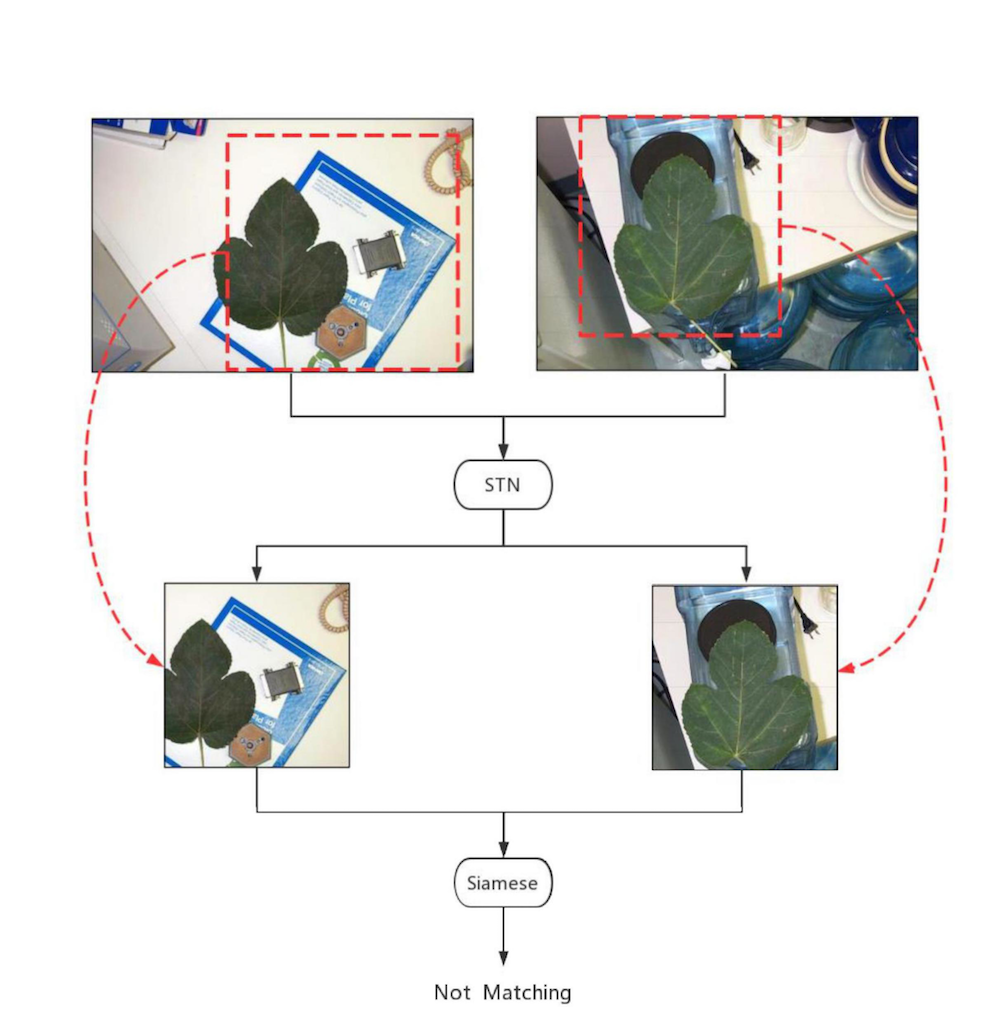}}
\subfigure[]{
\includegraphics[height=0.2\textwidth]{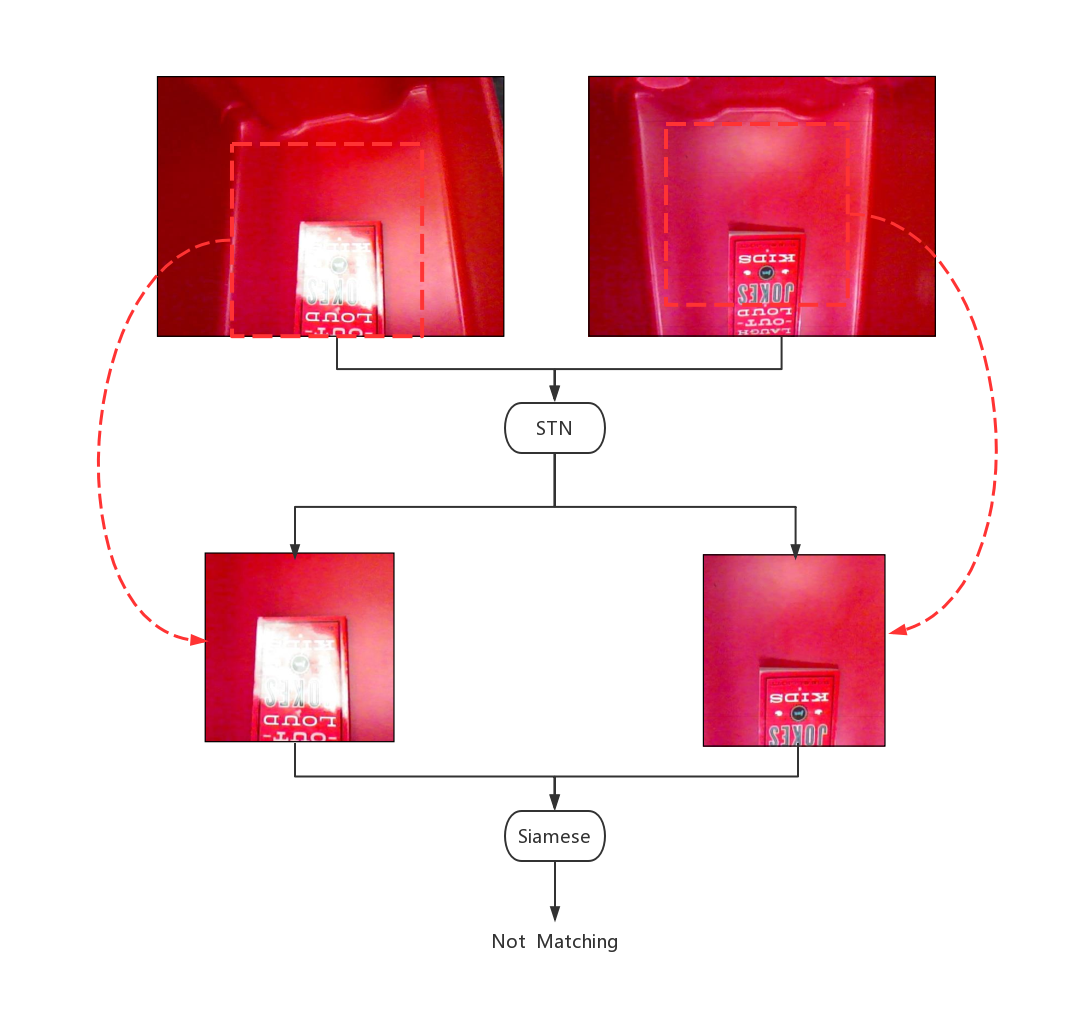}}
\caption{Two of examples that our method failed.}
\label{fig:Fig.9}
\end{figure}

Here are some examples that traditional Siamese network failed but our network succeeded (Fig.~\ref{fig:Fig.8}). And two of examples that our method failed are also presented as Fig.~\ref{fig:Fig.9}.

After analyzing the failed result, we can see that when the background of the image is complex which causes the STNs cannot detect the right object, it is more likely to perform badly. And the result can also be affected by the physical noises such as light and so on. But traditional method without STNs also cannot preform well either and our future work will work on it.

\subsection{Contribution}

In this paper, a generic object-centred image pairing method (i.e. determine whether an image pair contains the same object) is proposed that have promising results even the target object is small in the image or in a cluttered background. It  achieves this by a novel network structure named GetNet that combines Spatial Transformer Networks (STNs) and Siamese architecture.

Our idea is intuitionistic and reasonable. Humans have this amazing ability to home in on the parts of an image that contains the objects they are interested in even if the objects are inconspicuous. If retrieval systems that focus on the potential objects and ignore these ¡°distractions¡± just like what humans do can be build, and then apply similarity measurement to the potential objects in the image pairs instead of using whole-image similarity measurement, the accuracy and performance of image matching can be improved significantly.
In our approach, STNs are used to determine which parts of image in the query image pair to use for matching and outputs these subparts. Then Siamese architecture is applied to measure the similarity of the two subparts and determines whether they are pairing. Example pairs that is simply labelled "1"(means pairing) or "0"(means non-pairing) are given to train this network. 

Our approach not only improves the accuracy of image pairing problems, but also presents a new and effective method to train networks to get better performance. The alternately training method can be used to fully update the weights of the target part of network which is crucial for CNN to complete the task so that it can achieved better performance and it can be applied into other similar networks as well. At the same time, a new and efficient way for image ground truth auto-labeling is also provided since the target object can be located just given the label $1$ or $0$.

And also, the GetNet we proposed here provides a promising solution for finding lesion locations in medical image research. In recent years, the application of AI in the medical field has attracted the attention of more and more researchersa and medical image research is one of the most important aspects. The difficulty in medical image research is that images cannot be well understood as natural images, and doctors often have to determine the location of the lesion based on results such as cancer recurrence or lymph node metastasis, which can be very challenging. And our proposed method can help determine the region of interest and predict the outcome of the treatment, which we believe is instructive for future research.

\section{FUTURE WORK}
At present, the STN in our method only output two parameters,  which can only achieve the translation. In the next step, we will further study the STN and make it output more parameters to complete the rotation and other affine transformations to get better performance. At the same time the training of current network is difficult and the network structure will be optimized in order to simplify training process in the future. And the noises effect will also be overcome later on.

\section{CONCLUSION}
To solve the problem of pairing images contained the same objects, we propose a new network structure named GetNet with Spatial Transformer Network and Siamese network. Through the experiment, we confirm that our network can improve the accuracy and also label the target object effectively. In this paper, a new and efficient way is also proposed to train the CNNs and it can help improve the performance of the network in some tasks.

\bibliographystyle{IEEEtran}
\bibliography{1}

\end{document}